\documentclass[letterpaper]{article}
\usepackage{iccc}
\usepackage[english]{babel}
\usepackage{blindtext}

\usepackage{graphicx}
\usepackage{times}
\usepackage{helvet}
\usepackage{courier}

\usepackage[numbers,square]{natbib} 
\usepackage{multirow}
\usepackage{booktabs}

\usepackage{color}
\usepackage{url}
\usepackage{subcaption}
\usepackage{units}
\usepackage{paralist}
\usepackage{flushend}
\usepackage{nicefrac}

\title{A Rule-Based Computational Model of Cognitive Arithmetic}
\author{Ashis Pati, Kantwon Rogers and Hanqing Zhu\\
Georgia Institute of Technology\\
Atlanta, Georgia, United States 30318\\
\{ashis.pati,kantwonrogers,williamzhu\}@gatech.edu\\
}
\begin{document} 
\maketitle
\begin{abstract}
\begin{quote}
Cognitive arithmetic studies the mental processes used in solving math problems. This area of research explores the retrieval mechanisms and strategies used by people during a common cognitive task. Past research has shown that human performance in arithmetic operations is correlated to the numerical size of the problem.  Past research on cognitive arithmetic has pinpointed this trend to either retrieval strength, error checking, or strategy-based approaches when solving equations. This paper describes a rule-based computational model that performs the four major arithmetic operations (addition, subtraction, multiplication and division) on two operands. We then evaluated our model to probe its validity in representing the prevailing concepts observed in psychology experiments from the related works. The experiments specifically explore the problem size effect, an activation-based model for fact retrieval, backup strategies when retrieval fails, and finally optimization strategies when faced with large operands. From our experimental results, we concluded that our model’s response times were comparable to results observed when people performed similar tasks during psychology experiments. The fit of our model in reproducing these results and incorporating accuracy into our model are discussed.
\end{quote}
\end{abstract}

\section{Introduction}
\label{sec:intro}
Performing arithmetic operations is an elementary skill that most people learn and acquire at an early age and is used extensively in our day to day lives. Primary school students spend a significant time during their education practicing and honing this skill. Cognitive arithmetic can be defined as the study of the mental representations and cognitive processes which occur when people perform arithmetic in their heads. A very similar definition provided by Ashcraft in \cite{ashcraft1992cognitive} states that cognitive arithmetic is \textit{``concerned with the mental representation of number and arithmetic, and the processes and procedures that access and use this knowledge.''} Study of cognitive arithmetic promises to provide useful insights into the working of our brain. It is an important tool which bridges the theoretical and practical understanding of  memory  \cite{fendrich2014mental}. Understanding these underlying mental processes can also help in designing more effective school curriculum to teach arithmetic to students. Hence, this area of research has attracted a lot of interest among the cognitive science community, more so from cognitive psychologists who have performed extensive empirical studies involving human participants. These studies have been instrumental in unraveling several important phenomena associated with cognitive arithmetic and have attempted to provide several theories to explain them. 

Although the theories based on observations from psychological experiments are intuitive, they cannot be validated easily and sometimes it is difficult to reproduce and extend them. On the other hand, computational models are powerful tools in cognitive science to evaluate and test the validity of theories. Computational models are generally scalable allowing detailed experimentation with any number of inputs while simultaneously allowing precise control of test conditions. An accurate computational model can also be used to predict human behavior and hence, is extremely useful. For e.g. experiments after ``ablating'' the model can be used to predict the behavior of neuro-psychological patients. Hence, coming up with a computational model which explains the  theories is essential to any area of study in cognitive science. 

In this paper, we first provide a brief discussion on the previous research done in this area. We will explain several commonly observed phenomena in experimental studies on cognitive arithmetic and the theories proposed to explain them. We then draw ideas and data from these experiments to create computational models of cognitive  arithmetic. We create two basic models: a) A overall model for how cognitive arithmetic is performed, b) A secondary model which tries to explain how common arithmetic facts are stored in and retrieved from memory. Subsequently, we design a few experiments which test our model in an attempt to evaluate and ascertain their validity. Finally, based on the experimental results, we identify several shortcomings in our models and discuss possible directions for future research endeavors. 

\section{Related Work}
\label{sec:rel}
Although, cognitive arithmetic covers number theory and math problems, our work strictly deals with basic arithmetic equations. The body of existing research in cognitive arithmetic follows an approach grounded in cognitive psychology by employing empirical human-subject research and computational modeling to explain mathematic cognitive processing. Past empirical research test hypotheses regarding the cognitive mental processes and structures by examining participants’ response times, error rates and also by asking the participants to explain their thought process. The data thus obtained from participants solving arithmetic problems can be used to inform the results from computational models. While our approach did not include human-subjects research, we were able to leverage this past work and data to inform our model. In this section we discuss the empirical research considered for constructing our computational model.


\subsection{Psychology Studies}
Early research by Groen and Parkman in cognitive arithmetic involved studying the performance of adults completing addition problems \cite{groen1972chronometric}. Their work identified the phenomena of \textit{``tie-effect''} in cognitive arithmetic, wherein it was observed that people are faster at solving problems with the same operands, e.g. 4 + 4 = 8. Groen and Parkman later completed a study with first-grade children. The children in the study were asked to complete small addition problems (the sum did not exceed 9). The researchers found that as the sums increased in size, the response times of the children linearly increased as well. This led them to believe that children use a \textit{``minimum counting''} strategy i.e. counting begins from the larger of the two operands and increments for the value of the smaller number.

Other researchers at that time contested the counting-based model being the only mechanism for solving math problems. They instead proposed a hypothesis that adults use direct access or fact retrieval in order to solve arithmetic equations. Other researchers theorized that counting and direct access were only part of the cognitive model for solving arithmetic equations and that sequential, independent stages are used. Ashcraft inspected the theory of direct access and sequential stages by completing an addition study with adults \cite{ashcraft1978cognitive}. The study found a nonlinear increasing response time compared to the sum (when ties were removed), which supported Ashcraft’s hypothesis that counting could not be the only mechanism for solving these problems. Ashcraft and Stazyk later conducted tests of their data with the Groen and Parkman work to generalize the increasing response time in relation to the size of the operands as the \textit{``problem-size effect''}. They claimed that in general response time and error rates increase as the problem-size increases, where ``size'' refers to the sum of the addition problem. This finding has led to a large body of work aimed at understanding and documenting the response time and error rates of participants when completing arithmetic problems. 

In the Miller and Perlmutter work, the researchers compared the performance of adults in completing addition, multiplication and numerical comparison problems \cite{miller1984cognitive}. Miller and Perlmutter found the relationship of the response times to support a ``network'' or sequential model for computing the solutions as opposed to a counting model. The authors also found evidence for a distinction between location and accessibility of information within a direct access model. This means that fact recall could vary based on factors such as how often or recently a person has retrieved that arithmetic fact. Research by Campbell and Xue looked into the strategies and performance of people across cultures for all four major math operations \cite{campbell2001cognitive}. The study confirmed the importance of procedural knowledge when successfully completing arithmetic problems. This work correlates participants' skill in arithmetic to their ability to complete complex arithmetic procedures (e.g. borrowing, carrying, place holding), and not entirely based on retrieval performance. Researchers have also studied the processes used when recall fails participants. This area of research has hypothesized a number of backups strategies used to indirectly solve math problems, however a general consensus has not been reached due to the variety of responses found during the studies \cite{LeFevre1999,Geary1993,campbell2008subtraction}. This research has shown that people actively use backup strategies, such as counting by multiples, subtraction by addition, and operand transformation when they cannot directly access an answer.

\subsection{Computational Modeling}
Initial computational models to explain cognitive arithmetic were developed by Groen and Parkman using information processing models for counting \cite{groen1972chronometric}. As previously mentioned, the authors found that the response times generated by \textit{``minimum counting''}  best fit their sample data when compared to four other counting models. Ashcraft also accompanied their psychology study with a computational model based on direct access. Ashcraft developed a network retrieval model that inspected sequential stages of arithmetic processing in combination with fact retrieval \cite{ashcraft1992cognitive}. This work helped recognize that basic fact retrieval was used in conjunction with the carry operation process to solve more complex addition problems (e.g. two digit operations) and opened up research initiatives in network models based on the activation and priming of arithmetic facts through network models \cite{umilta2005computational,lebiere1999dynamics}. For example, Lebiere developed an ACT-R model of cognitive arithmetic based on the findings from the psychology experiments to simulate “a lifetime of arithmetic learning” \cite{lebiere1999dynamics}. Similar to our own model, Lebiere’s  uses an activation-based process to learn arithmetic facts. 

\section{Proposed Model}
\label{sec:model}

\begin{figure}[h]
  \includegraphics[width = 0.50\textwidth]{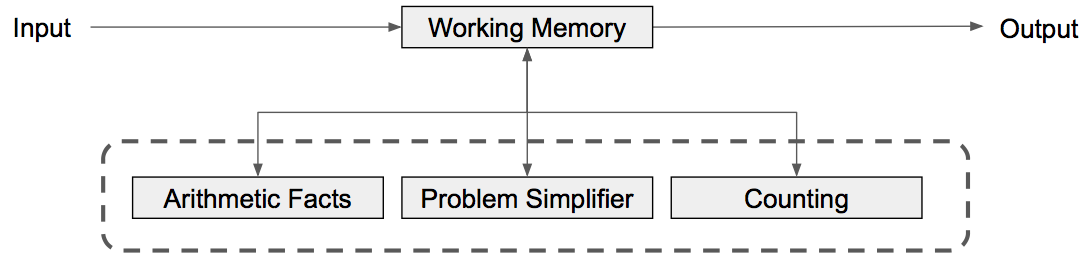}
    \caption{Overall Arithmetic Computation Model}
    \label{fig:overall_model}
\end{figure}

For the purpose of this paper, we designed a model which could solve elementary two-operand operations (addition, subtraction, multiplication and division). The overall model has three steps:  the input parsing, working memory operations, and output generation. Figure~\ref{fig:overall_model} show the overall system as a block digram. The input to the model is a string that contains the first operand then an operator followed by the second operand. Available operations are addition, subtraction, multiplication, and division. Once the input has been parsed, the working memory is tasked with the computation of the problem. The primary mode is memory retrieval. The working memory starts by trying to fetch the result of an operation from a database of entries we call the arithmetic fact library. This models how a person who has memorized an operation retrieves it from his or her memory. If this retrieval fails, the working memory then attempts to simplify the problem by breaking it down into smaller processes and continues to check if any parts of the simplified version are found in the database. Simplification processes are dependent on the operation and based on common algorithms that humans execute when performing arithmetic. If the problem can no longer be simplified and the operation is not available in the database, the system resorts to a counting method. Throughout this entire process, overall response time is calculated per operation based on reported response times of humans performing the respective arithmetic task. After an answer has been produced, the system displays the result and the overall response time.  

\begin{figure}[h]
\centering
  \includegraphics[width = 0.31\textwidth]{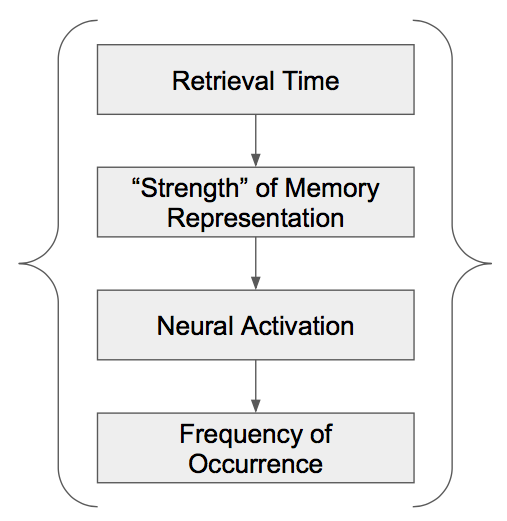}
    \caption{Model for Arithmetic Facts Retrieval}
    \label{fig:rt_model}
\end{figure}

For operands that exist in the arithmetic fact library, the total response time for such a problem is the RT stored in memory. Based on this assumption, the problem-size effect (where larger sums take longer time) can be explained only if we can come up with a computational model for these RTs. So we designed a computational model for the RTs associated with individual operand pairs using the neural activation model proposed in \cite{hamann1986textbook}. We conjecture that RT for an operand pair is inversely related to the strength of the memory representation associated with that operand pair. Assuming a neural model for memory representation, the strength of the representation associated with an operand pair would be directly proportional to the neural activation. Neural activation in return depends on the number of times the operand pair is presented to the user. In other words, RT of an operand pair is inversely proportional to the frequency of occurrence of that pair. The complete model for arithmetic fact RT is shown in Figure~\ref{fig:rt_model}. Since, we assume that basic arithmetic facts (operations on single digit operands) are stored in memory, the response time for those will be the same as memory RT. Thus, we conjecture that the experimental data reported by Miller et al. \cite{miller1984cognitive} for e.g. corresponds to the memory RT.

\section{Experiments}
\label{sec:exp}
To evaluate the representational power of our model, we conducted three experiments to compare its fit to the  findings of psychology experiments discussed in the related works. 

\subsection{Problem-Size Experiment}
\begin{figure}[h]
\centering
  \includegraphics[width = 0.40\textwidth]{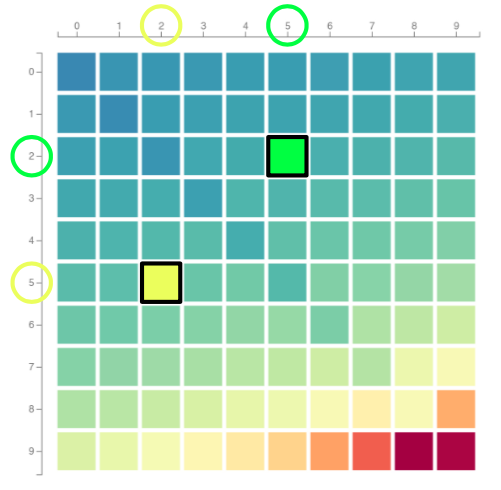}
    \caption{Activation modeling for an input pair (2,5)}
    \label{fig:activation_model}
\end{figure}

\begin{figure}[h]
  \begin{subfigure}[b]{0.23\textwidth}
    \includegraphics[width=\textwidth]{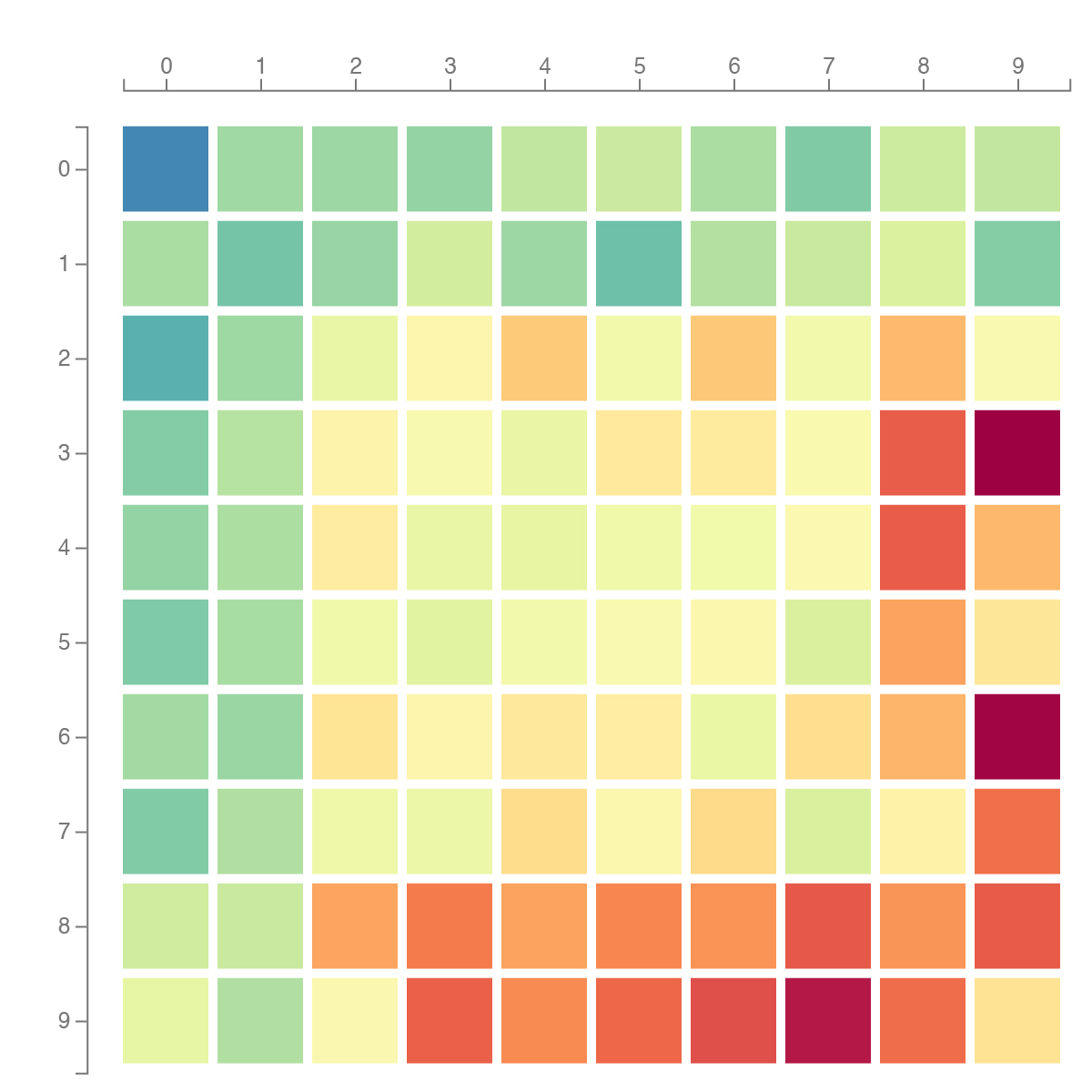}
    \caption{}
    \label{fig:exp_model}
  \end{subfigure}
  \begin{subfigure}[b]{0.23\textwidth}
    \includegraphics[width=\textwidth]{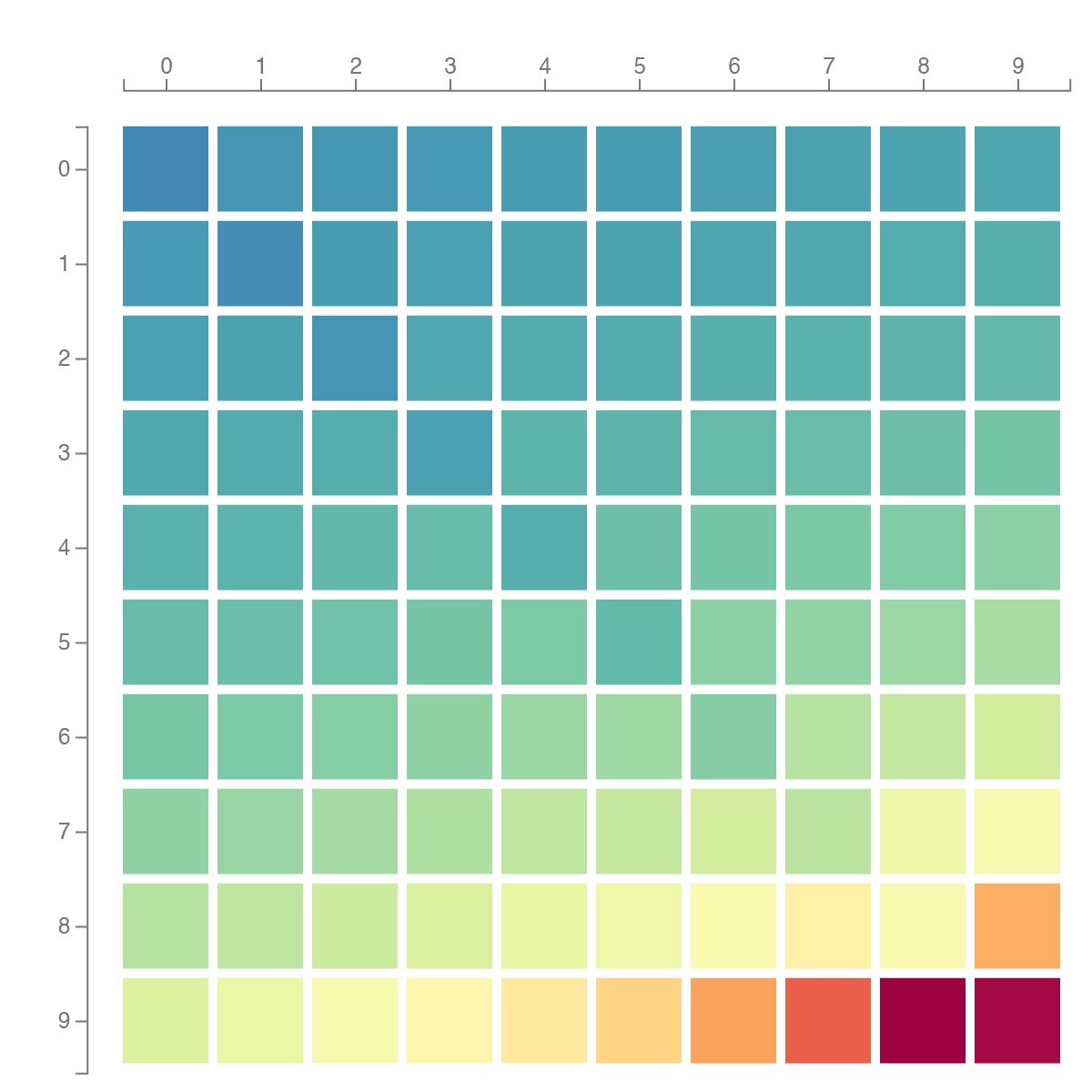}
    \caption{}
    \label{fig:our_model}
  \end{subfigure}
  \centering
  \begin{subfigure}[b]{0.25\textwidth}
    \includegraphics[width=\textwidth]{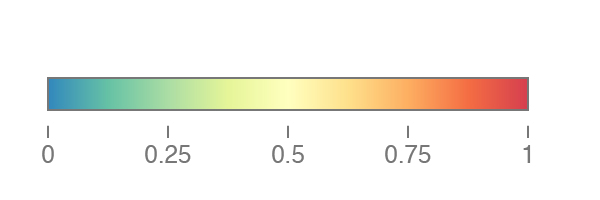}
  \end{subfigure}
  \caption{Normalized Retrieval Times: a) From experimental data provided by \cite{miller1984cognitive}, b) From our model simulation }
  \label{fig:size_effect}
\end{figure}

To test our model for arithmetic fact retrieval (Figure~\ref{fig:rt_model}) we designed an experiment to model the frequency of occurrence of a pair of numbers while learning multiplication. Since, most of elementary school students learn single-digit multiplication via the use of multiplication tables \cite{ashcraft2009mathematical}, we decided to use the same in order to determine the frequency of occurrence of a number pair. We assumed that learning of multiplication tables is a pseudo-random process wherein for ``trial'', the table being memorized (e.g. table of 2 corresponds to 2x0, 2x1,....till 2x9) is selected randomly. We call this the table index. Also, every time the entire table is not memorized and the extent to which a table is memorized is also random. For example, when learning multiplication by2, in one instance we go from 2x0, 2x1,....till 2x8 and in another instance we go from 2x0, 2x1,....till 2x5 only. In these examples 8 and 5 would be the table extent. Next, we initialized a random number pair (each number is between 0 to 9 drawn from a uniform distribution). The first number corresponds to the the table index and the second number corresponds to the table extent. For e.g. if the we get a pair (2,5), this means we memorize 2x0, 2x1, .... till 2x5 for this trial. Therefore, in our memory representation corresponding to multiplication, the locations corresponding to (2,0), (2,1),...till (2,5) get activated. We also conjecture that the conjugates of these locations i.e. (0,2), (1,2),....till (5,2) also get activated but to a lessor degree as shown in Figure~\ref{fig:activation_model}. The learning factor that we use for conjugate location activation in our experiment is 0.5. This trial is repeated a large number of times (10000). The resulting activation for a number pair is indicative of the strength of it's memory representation and the inverse of this models the RT. The RTs of our model and the response times from the experiments conducted by Miller et al. \cite{miller1984cognitive} are shown in Figure~\ref{fig:size_effect}. Note that both matrices are normalized between 0 and 1. 

\subsubsection{Discussion}
At first glance, the two plots shown in Figure~\ref{fig:size_effect} appear to be different. However, a closer look provides some useful insights:
\begin{enumerate}
\item The problem-size effect is clearly visible in both plots. As the multiplicands increase, there is an increase in the corresponding RTs. The reason this happens in our simulation (Figure~\ref{fig:our_model}) is because in our experimental design, the frequency of occurrence of smaller numbers is much higher than that of larger numbers while memorizing multiplication tables. Hence, problem-size effect might have its roots in the way arithmetic operations are memorized and practiced. This conclusion from our simulation model also validates some of the experiments reported by Hamann \& Ashkraft \cite{hamann1986textbook} in which they they looked at the addition problems in textbooks from three publishers for grades K through 3. They also found out that the frequency of operands 2,3 or 4 was significantly higher than operands 5-9. Similar results were also reported by Clapp \& Heubner \cite{clapp1924number}.
\item The tie-effect wherein the RT for an operand pair with two identical digits is comparatively lower than its surrounding pairs on the plot is also observed in both plots, albeit to a greater degree in our simulation model. This can be seen as the slightly ``greener'' diagonal line running from the top-left to the bottom-right corner of Figure~\ref{fig:our_model}.
\end{enumerate}
Hence, even with significant differences our simulation model is able to account for two important effects which are observed in experimental data which supports our assumption that for single digit operands, we employ a memory based retrieval process. In the subsequent experiments we thus use the response times reported by Miller et al. \cite{miller1984cognitive} to model our RTs.

\subsection{Ablation Experiment}

Often, to learn about how a working system operates, one can remove a component and study how the system compensates, a process called as ablation. Since our cognitive arithmetic system depends on the retrieval of archived operations, we removed a subset of these operations and determined the effect on response time. Based on prior research, novices resort to a counting strategy when exposed to an addition or subtraction problem in which they do not have operations memorized. Therefore, we incorporated this functionality into our system. Additionally, research by Ashcraft \& Guillaume \cite{ashcraft2009mathematical} suggests that novices commonly forget single-digit operations involving the numbers 5,6,7,8, and 9. Thus, we chose to ablate all operations where both operands contained either of these numbers and then measure the response times. 

\subsubsection{Discussion}

\begin{figure*}[h]
  \begin{subfigure}[h]{0.45\textwidth}
    \includegraphics[width=\textwidth]{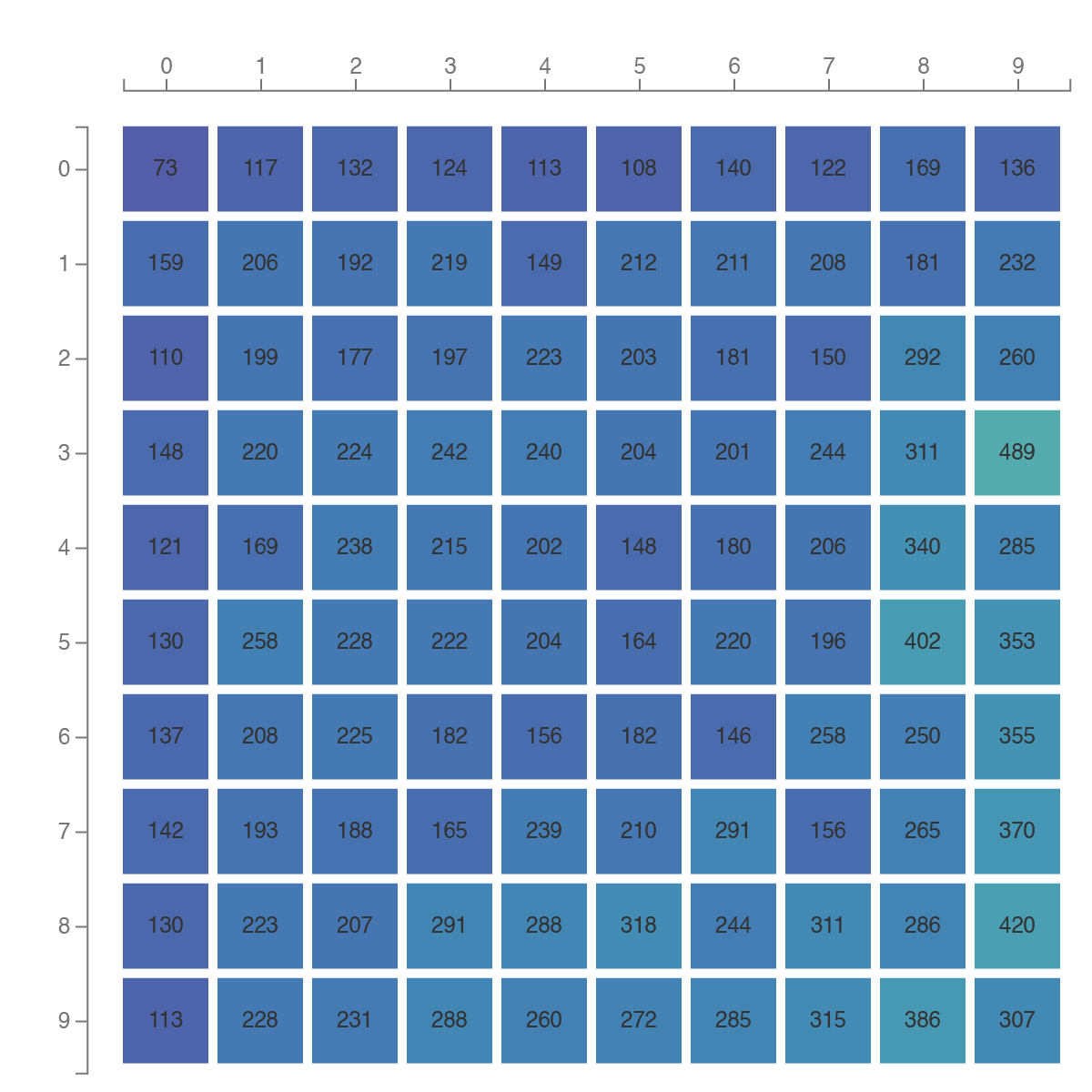}
    \caption{}
    \label{fig:no_ablation}
  \end{subfigure}
  \begin{subfigure}[h]{0.45\textwidth}
    \includegraphics[width=\textwidth]{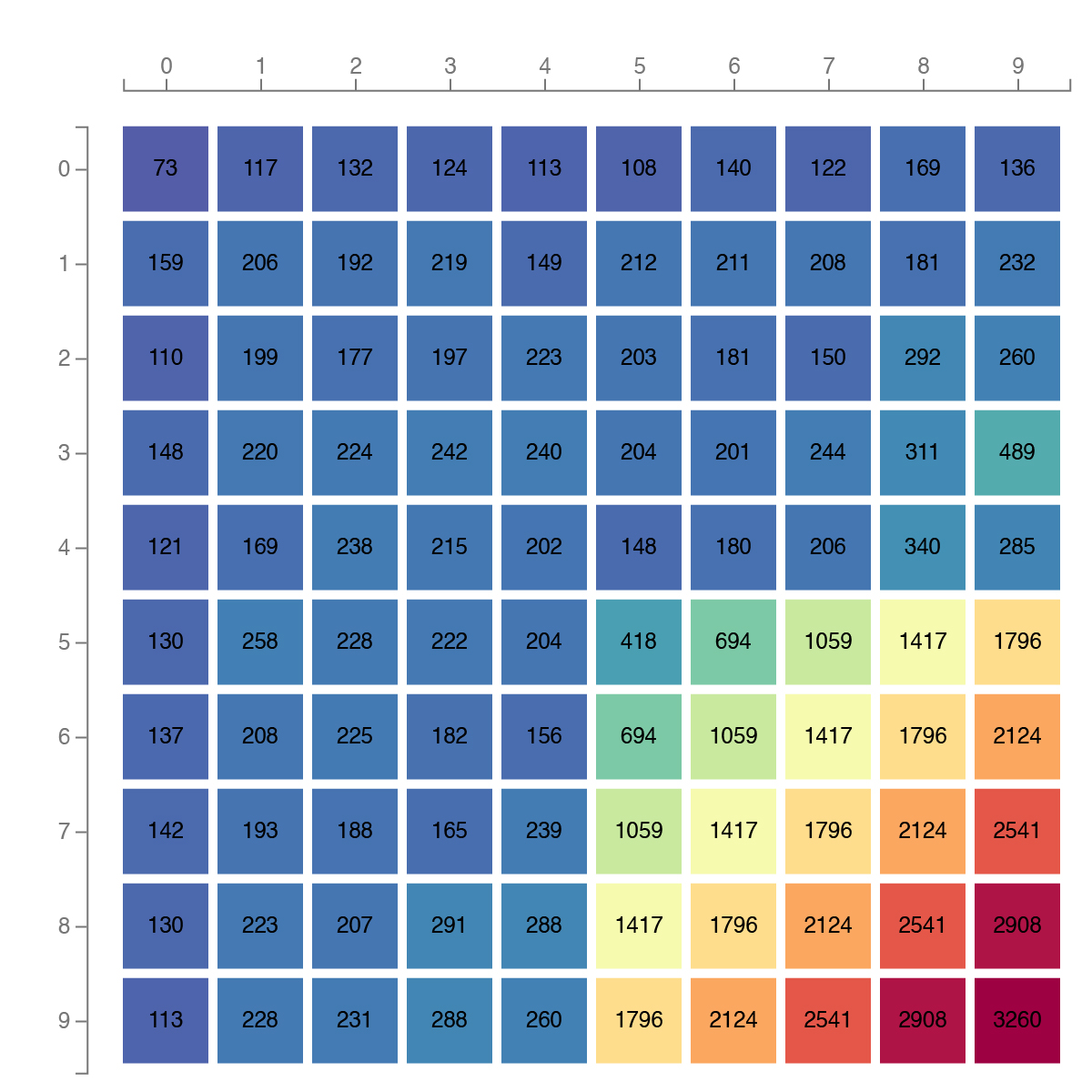}
    \caption{}
    \label{fig:ablation}
  \end{subfigure}
  \centering
    \begin{subfigure}[h]{0.3\textwidth}
    \includegraphics[width=\textwidth]{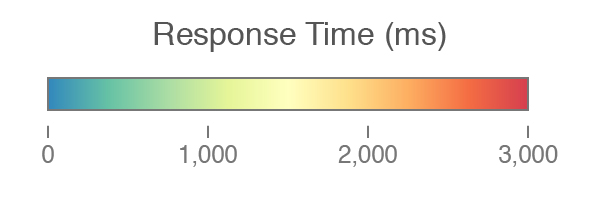}
  \end{subfigure}
  \caption{Response Times in Milliseconds a) Without Ablation b) With Ablation}
 \label{fig:ablationExperiment}
 
\end{figure*}

Figure~\ref{fig:ablationExperiment} shows the response times for all single-digit addition operations with and without ablation. It is evident that ablation and the resulting counting strategy results in significantly higher response times. This provides further evidence that counting based strategies are not generally employed for operations involving single digit operands because if that were the case the response times observed in experimental data would resemble Figure \ref{fig:ablation} more closely.

\subsection{Strategy Optimization Experiment}
Because of the problem-size effect wherein problems with larger digit values tend to have a longer RT, it is possible for one to create to optimized strategies for certain types of problem from a metacognition perspective. In other words, it might be beneficial to transform a typical problem into a problem with operands which have smaller digit values. 

Needless to say, it usually takes one huge amount of practice to find out a better strategy for a certain type of problem. In this process, a person uses logic to find out these strategies with several attempts to the problem. Then, analogy is employed so that the strategy can be applied to a similar problem. One example of such strategy is ``Division by 5'', where, in certain cases, it is more beneficial to multiply the numerator and the denominator both by 2 first, which results the denominator to be 10. For example, for $\frac{1375}{5}$, it will be better to implement a strategy so that $\frac{1375}{5} = \frac{1375*2}{5*2} = \frac{2750}{10} = 275 $. Another common strategy is ``Fast Addition'', wherein whenever an operand is close to a number with lots of zeros at the end (100, 1000, 5000, 40000 etc.), it may be faster to first round the operand to that number, subtract the difference from the second operand and then add the two new operands. For example, for $497 + 38$, it will be faster to have $497 + 38 = (497+3) + (38 - 3) = 500 + 35 = 535$.

\begin{figure}[h]
\centering
  \includegraphics[width = 0.48\textwidth]{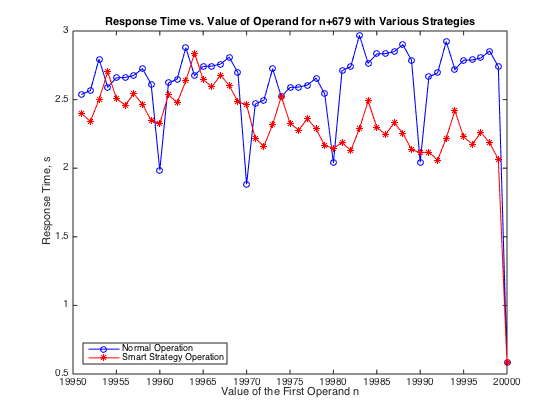}
    \caption{Response Time vs. Value of Operand for n+679.}
    \label{fig:smart50}
\end{figure}

We tried to simulate this strategy in this experiment. Consider $n + 679$, where $n$ is from 19950 to 20000. The response time from our model for that operation with and without Fast Addition strategy is plotted in Figure~\ref{fig:smart50}. 

\subsubsection{Discussion}
It is clear to see that the larger the distance of the augend is from 20000, the lower the difference between the response times of the two strategies. For cases where the augend is far away from 20000, it is sometimes not worth it to implement Fast Addition at all (red cure above blue curve). Take $19992 + 679$ as one example. The response time from our model for that operation is 2.694 seconds without using Fast Addition and 2.059 seconds with Fast Addition. On the other hand, for $19952 + 679$, it takes 2.566 seconds without using Fast Addition and 2.341 seconds with Fast Addition. The results can been seen in Table~\ref{smartTable}.

\begin{table}[h]
\centering
	\caption{Comparison between Operations for Fast Addition.}
	\label{smartTable}
    \begin{tabular}{|c|c|c|}
    \hline
    Sum & Distance: n to 20000 & Time Saved (s)\\ \hline
    19992 + 679 & 8 & 0.635 \\ \hline
    19952 + 679 &  48 & 0.225 \\ \hline
    \end{tabular}
\end{table}

\begin{figure}[h]
\centering
  \includegraphics[width = 0.48\textwidth]{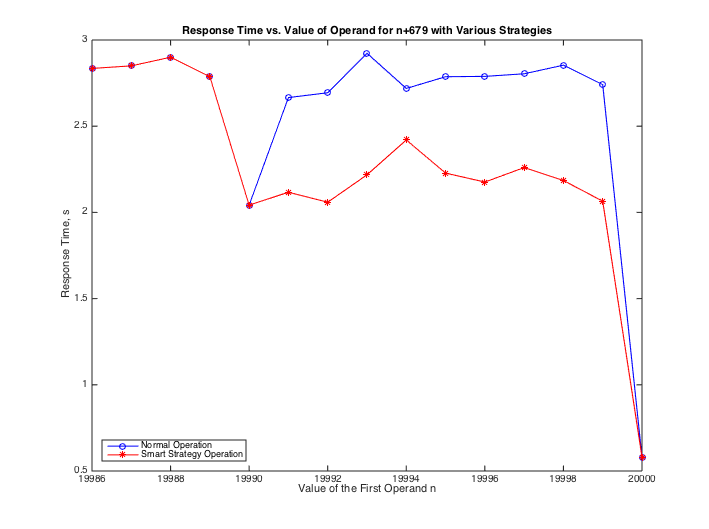}
    \caption{Response Time vs. Value of Operand for n+679 with the Optimized Strategy.}
    \label{fig:smart10}
\end{figure}

It can be seen in Table \ref{smartTable} that the time saved by using Fast Addition is 0.635 seconds for $19992 + 679$ and 0.225 seconds for $19952 + 679$. The reason for this phenomenon is that if the distance is a 2-digit number with a large value, it might not be worth it at all to implement the Fast Addition strategy since it will not save a lot of time. If we consider accuracy as well, it is more likely to make mistakes in operations with 2-digit numbers than with 1-digit numbers. Thus, an optimal threshold for the Fast Addition strategy might be a distance within 10 to a number with more zeros at the end (20000 in this case).  For example in Figure~\ref{fig:smart10}, the actual improvement in response time is shown when the Fast Addition strategy is employed from 19991.

In general, there exist a large number of time saving strategies for all operations. Using logic from several attempts to a specific type of problems, it is possible to find out a pattern and solve the problem with minimal time. Furthermore, the same strategy can be applied to similar problems with certain patterns by analogy.  It is also interesting to note that the reason these optimizing strategies work is because of the problem-size effect which in turn might be  the result of the non-uniform way in which we learn arithmetic.

\section{Conclusion \& Future Work}
\label{sec:conc}
In this study, a rule-based computational model is proposed for cognitive arithmetic, where frequent memory updates and retrievals for arithmetic rules and facts are needed. From the experiments performed, it is clear to see the RT is strongly correlated to how frequently a specific operand pair is encountered. The more frequently an operand pair appears, the faster one is able to respond to it. In order to validate the rule-based model, an ablation experiment was performed. The ablation of certain arithmetic rules and facts resulted in a dramatic degradation to the overall performance of the model, where, in this case, the model resorted to counting, which is the fundamental method in arithmetic operations. It is also shown that response time depends on the problem size. A problem with larger numbers tends to result in a slower response. Using this, it is possible for one to find out better strategies that can to be employed for certain operations. In other words, it becomes possible to optimize the response time using a strategy, where operations with larger numbers are avoided. In our rule-based computational model, the optimization strategies for certain operations are successful and perform as expected.

A significant drawback of our model is that we don't have any way to model the accuracy of the system. Our model is always accurate and that is one area where it differs from an actual human. The future work based on this model is to take accuracy into consideration in modeling cognitive arithmetic. Accuracy is another important evaluation parameter besides response time for cognitive arithmetic. The relationship between accuracy and problem size can be studied. Furthermore, the relationship between response time and accuracy can be determined since typically the faster a person responds to a problem the more likely he will make a mistake. Finally, the optimization over strategies can be upgraded to the next level where the new target is to not only produce a fast but also an accurate response to any arithmetic operation.   

\bibliographystyle{iccc}
\bibliography{iccc}

\end{document}